%% file: 2016_qcri_semeval3.tex
\documentclass{sig-alternate-05-2015}

\usepackage{times}
\usepackage{latexsym}

\usepackage{url}
\usepackage{color}
\usepackage{amsmath} 
\usepackage{tabularx}
\usepackage{chato-notes}

\newcommand{\Ni}{({\em i})~}
\newcommand{\Nii}{({\em ii})~}
\newcommand{\Niii}{({\em iii})~}

\newcommand{\Na}{({\em A})~}
\newcommand{\Nb}{({\em B})~}
\newcommand{\Nc}{({\em C})~}

\newcommand{\good}{\texttt{good}\,}

\newcommand{\perf}{\texttt{perfect match}\,}
\newcommand{\rel}{\texttt{relevant}\,}
\newcommand{\irel}{\texttt{irrelevant}\,}

\newcommand{\dir}{\texttt{direct}\,}

\title{%
Addressing Community Question Answering \\in English and Arabic
}

\author{Giovanni Da San Martino$^{\dagger}$, Alberto Barr\'on-Cede\~no$^{\dagger}$, \textbf{Salvatore Romeo$^{\dagger}$,}
\textbf{Alessandro Moschitti$^{\dagger}$,}
\textbf{Shafiq Joty$^{\dagger}$},
\textbf{Fahad A. Al Obaidli$^{\dagger}$,} \\
 \textbf{Kateryna Tymoshenko$^{\ddagger}$,}
 $^{\dagger}$ALT group, Qatar Computing Research Institute, Hamad Bin Khalifa University, Qatar\\
 $^{\ddagger}$Department of Computer Science and Information Engineering, University of Trento, Italy\\
  {\tt \{gmartino,albarron,sromeo, amoschitti, sjoty,faalobaidli,\}@qf.org.qa}\\
  {\tt \{kateryna.tymoshenko, antonio.uva, d.bonadiman\}@unitn.it} \\
}

\date{}

\begin{document}

\conferenceinfo{SIGIR'16}{July 17-21, 2016, Pisa, Italy.}
\CopyrightYear{2016} 
\setcopyright{rightsretained}
\conferenceinfo{SIGIR '16,}{July 17 - 21, 2016, Pisa, Italy}
\isbn{}
\doi{}
\clubpenalty=10000 
\widowpenalty = 10000

\numberofauthors{2} 
%
\author{
%
%
\alignauthor
Giovanni Da San Martino, Alberto Barr\'on-Cede\~no, Salvatore Romeo,  
 Alessandro Moschitti, Shafiq Joty, Fahad A. Al Obaidli \\
  \affaddr{Qatar Computing Research Institute}\\
  \affaddr{HBKU, Doha, Qatar}\\
  \email{\{albarron,gmartino,sjoty, faalobaidli, amoschitti,sromeo\}@qf.org.qa} 
\alignauthor
Kateryna Tymoshenko, Antonio Uva, \\ Daniele Bonadiman\\ 
       \affaddr{Department of Computer Science and Information Engineering, }\\
       \affaddr{University of Trento}\\
       \affaddr{Trento, Italy}\\
       \email{\{kateryna.tymoshenko, antonio.uva, d.bonadiman\}@unitn.it}
}

\maketitle
\input{abstractAndIntro}

\section{Related Work}
\label{sec:related}

The first step for any system that aims to automatically answer questions on cQA sites is to retrieve a set of questions similar to the user's input. The set of similar questions is later used to extract possible answers for the input question.  However, determining question similarity remains one of the main challenges in cQA due to problems such as the "lexical gap".
To overcome this problem, different approaches have been proposed.
Early methods used statistical machine translation techniques to compute semantic similarity between two questions.
For instance, \cite{zhou2011phrase} applied a phrase-based translation model. 
Their experiments on Yahoo! Answers showed that models based on phrases are more effective than those using words, as they are able to capture contextual information. 
However, approaches based on SMT have the problem of requiring lots of data in order to estimate parameters.

Algorithms that try to go beyond simple text representation are presented 
in~\cite{cao2009use} and~\cite{duan2008searching}. In~\cite{cao2009use} a 
similarity between two questions on Yahoo!~Answers is computed by using a 
language model with a smoothing method based on the category structure of Yahoo! 
Answers. In~\cite{duan2008searching}, the authors search for semantically 
similar questions by identifying the topic and focus of the user's question. 
More specifically, they compute a similarity between the questions' topic, 
which represents general users interests, and the questions' focus.
A different approach using topic modeling for question retrieval was introduced 
in~\cite{ji2012question} and~\cite{zhang2014question}. Here, the authors use 
LDA topic modeling to learn the latent semantic topics that generate 
question/answer pairs and use the learned topics distribution to retrieve 
similar historical questions. 
The quality of the ranking returned by all these systems was  measured on a set of test questions from Yahoo! Answers, with question relevancy judgment annotated by users, sometimes assigned automatically based on heuristics.

It should be noted that the methods above exploited language models or  general 
knowledge given by Yahoo! Answers~categories or LDA topics, whereas in our 
paper, we model the syntactic/semantic relations between pairs of questions 
using shallow syntactic parsing and lexical matching. 
The most similar work to ours is~\cite{wang2009syntactic}, where the authors 
found semantically related questions by computing the similarity between the 
syntactic trees of the two questions. They used a tree similarity computed in 
terms of the number of substructures shared between two trees.
%
Different from such approach, we use pairs of questions, $(q_{o}, q_{s})$, as 
learning instances, thus defining relational models connecting the syntactic 
trees of $q_{o}$ and $q_{s}$. In this way, the learning algorithms learn 
transformations that suggest if questions constituted by similar words have 
similar (paraphrases) or different semantics.

\section{Problem Description} 
\label{sec:taskdescription}

Among the four task of the SemEval-2016 competition, we focus our attention on 
the two dealing with question--question similarity. In task B English we 
consider the questions' text only. In task D Arabic we also use the information 
from the answer linked to the forum question.
Tasks B uses English instances extracted from \textit{Qatar Living}, a forum for people to pose questions on multiple aspects of daily life in Qatar.%
\footnote{\url{http://www.qatarliving.com/forum}}
Task D uses Arabic instances extracted from three medical fora: \textit{webteb}, \textit{altibbi}, and \textit{consult islamweb}.%
\footnote{\url{https://www.webteb.com/}, \url{http://www.altibbi.com/}, and \url{http://consult.islamweb.net}.} 
As these are re-ranking tasks, we evaluate our models using mean average 
precision (MAP); which is the official evaluation measure of the SemEval 
2016 task.

\subsection{Task B: English Question--Question Similarity }

\begin{table}
\caption{A re-ranking example for the English Question--Question Similarity 
dataset. For each candidate the Google rank (G), the binary gold standard (GS) 
relevance, and our rank (R) are reported.}
\label{tab:example}
\centering
\tabcolsep=0.05cm
\begin{tabularx}{\columnwidth}{c|c|c|c}
\multicolumn{4}{p{8.3cm}}{\textbf{$q_o$}: What are the tourist 
places in Qatar? I'm likely to travel in the month of June. Just wanna know some 
good places to visit.
}	\vspace{.2em}\\ \hline
\textit{G}	& \textit{GS} & \textit{R}	& \textit{Question Text}	
\\ \hline
1 & -1 & 8 & \multicolumn{1}{p{6.9cm}}{\footnotesize The Qatar banana island 
will be transfered by the end of 2013 to 5 stars resort called Anantara. Has 
anyone seen this island? Where is it? Is it near to Corniche?}
\\
2 & +1 & 2 & \multicolumn{1}{p{6.9cm}}{\footnotesize Is there a good place here 
where I can spend some quality time with my friends?}
 \\
3 & -1 & 7 & \multicolumn{1}{p{6.9cm}}{\footnotesize Where is the best beach in 
Qatar? Maybe a silent and romantic bay? Where to go for it?}\\
4 & -1 & 9 & \multicolumn{1}{p{6.9cm}}{\footnotesize Any suggestions on what 
are the happenings in Qatar on Holidays? Something new and exciting suggestions 
please?} \\
5 & -1 & 3 & \multicolumn{1}{p{6.9cm}}{\footnotesize Where in Qatar is the best 
place for Snorkeling? I'm planning to go out next friday but don't know where to 
go.} \\
6 & -1 & 6 & \multicolumn{1}{p{6.9cm}}{\footnotesize Can you give me some nice 
places to go or fun things to do in Doha for children 17-18 years old? Where can 
we do some watersports (just for once, not as a member), or some quad driving? 
Let me know please. Thanks.} \\
7 & +1 & 1 & \multicolumn{1}{p{6.9cm}}{\footnotesize Which all places are there 
for tourists to Qatar? My nephew 18 years on visit.}
 \\
8 & -1 & 10 & \multicolumn{1}{p{6.9cm}}{\footnotesize Could you suggest the 
best holiday destination in the world?} \\
9 & -1 & 5 & \multicolumn{1}{p{6.9cm}}{\footnotesize I really would like to 
know where the best place to catch fish here in Qatar is. But of course from 
the beach. I go every week to Umsaeed but rerly i catch somthing! So 
experianced people your reply will be appreciated.} \\ \hline
\end{tabularx}
\end{table}

Conceptually, question retrieval is not much different from a standard retrieval 
task. Given the asked (original) question $q_{o}$, a search engine seeks the Web 
(or a specific Web forum) for relevant webpages. In cQA, webpages are threads 
containing questions $q_{s}\in Q$, with their user comments, where the latter 
can provide information for answering $q_{o}$. For example, 
Table~\ref{tab:example} shows an original question followed by some questions 
retrieved by a search engine. In the SemEval dataset we used in our experiments, 
each question is associated with $10$ related threads that are the top 10 
webpages retrieved by Google. The main difference with standard document 
retrieval is the document scoring function. Indeed, although both question and 
comments are part of the candidate webpage (thread), the question text provides 
a more synthetic and precise information to infer whether the candidate thread 
is relevant for $q_{o}$. Table~\ref{tab:classdist} gives class-distribution 
statistics of the English dataset~\cite{nakov-EtAl:2016:SemEval}. A forum 
question can be a \perf, \rel, or \irel with respect to the new 
question. For evaluation purposes, both \perf and \rel instances are 
considered \rel and must be ranked on top of the \texttt{irrelevant}  
questions. The corpus is composed of $387$ user questions, each of which 
includes $10$ potentially related questions. The task organizers used the  
Google search engine, which represents also the strong baseline for the task, to 
select potentially relevant forum questions. 
Table~\ref{tab:relevant_distribution} shows the distribution of 
relevant/irrelevant forum questions per ranking position. Although relevant 
questions tend to be concentrated towards the top of the Google Rank, they are 
fairly spread over the entire ranking. 

\begin{table}[t]
\vspace{-1em}
\caption{Class distribution in the training, development, and test 
partitions for the English Question--Question similarity 
dataset.\label{tab:classdist}}
\centering
\footnotesize
\begin{tabular}{l|crrr}
Class	    & train	& dev 	& test	& overall	\\\hline
Relevant    & 1,083	& 214	& 233	& 1,530	\\
Irrelevant  & 1,586	& 286	& 467	& 2,339	\\
Total  	    & 2,669	& 500	& 700	& 3,869	\\ \hline
\end{tabular}
\end{table}

\begin{table}[t]
\caption{ Distribution of Relevant and Irrelevant comments at 
different R ranking positions of GR for the Question--Question similarity 
dataset.}
\label{tab:relevant_distribution}
\centering
\footnotesize
\begin{tabular}{r|cccc}
R	& train		& dev 		& test		& overall	\\\hline
1	& $0.21\pm0.05$	& $0.24\pm0.07$	& $0.40\pm0.11$	& $0.25\pm0.07$	\\
2	& $0.14\pm0.03$	& $0.18\pm0.02$	& $0.12\pm0.02$	& $0.14\pm0.03$	\\
3  	& $0.11\pm0.02$	& $0.10\pm0.01$	& $0.08\pm0.01$	& $0.10\pm0.02$	\\
4  	& $0.12\pm0.03$	& $0.08\pm0.01$	& $0.10\pm0.03$	& $0.11\pm0.03$	\\
5  	& $0.09\pm0.02$	& $0.09\pm0.01$	& $0.08\pm0.02$	& $0.09\pm0.02$	\\
6	& $0.08\pm0.02$	& $0.09\pm0.02$	& $0.05\pm0.01$	& $0.08\pm0.02$	\\
7	& $0.08\pm0.02$	& $0.07\pm0.01$	& $0.05\pm0.01$	& $0.07\pm0.02$	\\
8	& $0.06\pm0.01$	& $0.04\pm0.01$	& $0.03\pm0.00$	& $0.05\pm0.01$	\\
9	& $0.07\pm0.02$	& $0.06\pm0.01$	& $0.04\pm0.01$	& $0.07\pm0.02$	\\
10	& $0.05\pm0.01$	& $0.05\pm0.01$	& $0.04\pm0.01$	& $0.05\pm0.01$	\\ 
\hline
\end{tabular}
\vspace{-1em}
\end{table}

\subsection{Task D: Arabic Question--Comment Pairs Re-Ranking} 

\begin{table}[t]
\vspace{-1em}
\caption{Class distribution in the training, development, and test 
partitions for the Arabic Question--Comment pairs
dataset.\label{tab:classdista}}
\centering
\footnotesize
\begin{tabular}{l|crrr}
Class	    	& train		& dev 	& test	& overall	\\\hline
\dir+\rel    	& 18,329	& 7634	& 7619	& 33582	\\
\irel  		& 12,082	& 5868	& 5951	& 23901	\\
Total  	    	& 30,411	& 13502	& 13570	& 57483	\\ \hline
\end{tabular}
\end{table}

A new question and a set of thirty forum question--comment pairs are provided 
---the comment is always a correct answer to the forum question. 
The task consists in re-ranking the question--comment pairs according to three 
classes:
\Ni \dir: if it is a direct answer to the new question;
\Nii \rel: if it is not a direct answer to the question but it provides information related to 
the topic; and 
\Niii \irel: if it is an answer to another question, not related to the topic. 
For evaluation purposes, both \dir and \rel forum questions are considered as \good. 
Table~\ref{tab:classdista} reports statics on the class distribution of the data. 
There are about 60\% of forum questions classified as \good, thus the dataset is not very unbalanced. 

As in the case of English, the Arabic collection questions are user-generated. 
Since the latter was extracted from medical-domain fora, they show specific 
challenges: a mix of medical terminology (used by the physicians) and colloquial 
language (used by the patients). The texts are usually long (on average 
questions are $50$ and comments are $120$ words long). These characteristics 
pose specific challenges to some of the models which are effective for task B on 
English, i.e. the tree kernels. 

%
\section{Our L2R Models}
\label{sec:features}

The ranking function for both tasks can be implemented by a scoring function $r:{ Q} \times { Q} \rightarrow \mathbb{R} $, where ${ Q}$ is the set of questions. 
The function $r$ can be linear: $r(q_{o},q_{s})$=$\vec{w}\cdot \phi(q_{o},q_{s})$, where $\vec{w}$ is a linear model and $\phi()$ provides a feature vector representation of the pair, $q_{o},q_{s}$.


We adopt binary SVMs to learn $r$ from examples and ee model $\phi(q_{o},q_{s})$ with different feature sets, which we describe 
in Sections~\ref{sec:proposedapproaches} and \ref{sec:proposedapproachesf}. 

\subsection{Tree Kernel Models} \label{sec:proposedapproaches}
\label{TK}

\begin{figure*}[t]
\center
\includegraphics[scale=0.4]{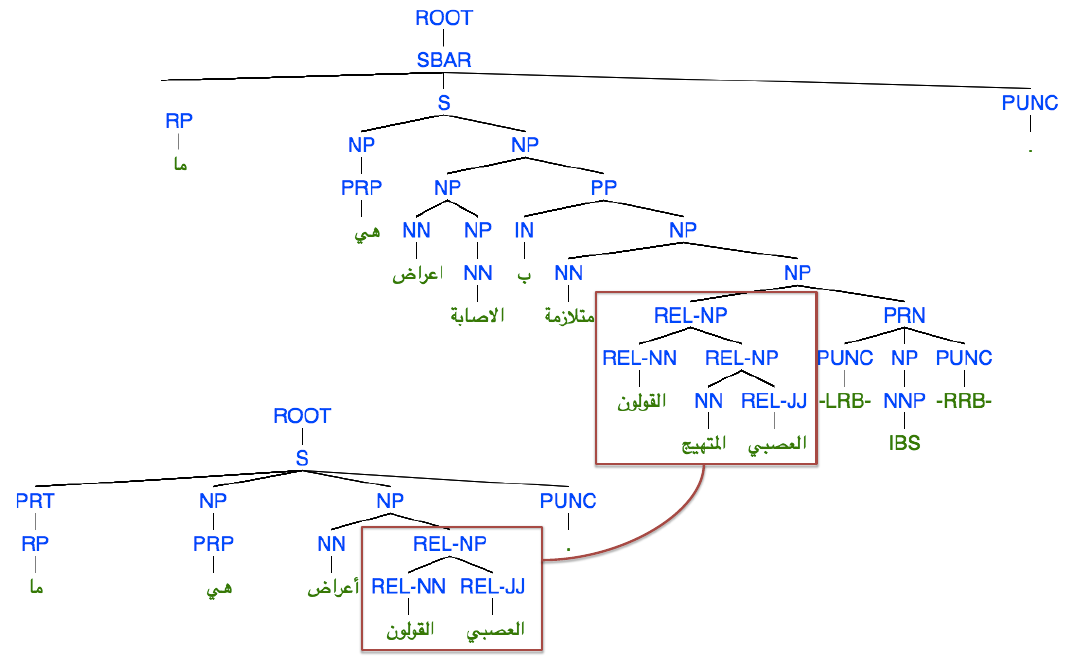} 
\caption{Our representation based on syntactic trees for the $q_o$--$q_s$ 
  pairs enriched with REL links. The translations are $q_o$ (sentence on top) ``What are the 
  symptoms of irritable bowel syndrome (IBS)?'' and $q_s$ ``What are the 
symptoms of irritable bowel?''.}
\label{trees}
\end{figure*}

Tree kernels are functions that measure the similarity between tree structures. 
We essentially used the model of~\cite{sigir12}, originally proposed to rank 
passages. Different from~\cite{sigir12}, our questions may contain multiple 
subquestions, a subject, greetings,  and elaborations, thus they are composed of 
several sentences.  We merge the whole question text in a macro-tree using a 
fake root node connecting the parse trees of all the sentences. In both tasks we 
represent pairs of questions; therefore,  we connect the constituents of two 
macro-trees corresponding to $(q_{o},q_{s})$, respectively. Figure~\ref{trees} 
shows an example where the relations between two Arabic questions are exploited 
to build a graph. The translation of the two questions is: 

\begin{description}
\item[$q_o$] {\it ``What are the symptoms of irritable bowel syndrome (IBS)?''}
\item[$q_s$] {\it ``What are the symptoms of irritable bowel?''}
\end{description}


\noindent We link the two macro-trees by connecting phrases, e.g., NP, VP, PP, 
when there is at least lexical matching between the phrases of $q_{o}$ and 
$q_{s}$. Note that such links are marked with the presence of a REL tag. 
Finally, we apply either a partial tree kernel (PTK) or the syntactic tree 
kernels (STK)~\cite{Moschitti:2006} and obtain the following kernel: 
\begin{align}
 K((q_{o},q^{i}_{s}),(q_{o},q^{j}_{s}) ) = & TK(t(q_{o},q^{i}_{s}),t(q_{o},q^{j}_{s})) \nonumber
 \\  &+ TK(t(q^{i}_{s},q_{o}),t(q^{j}_{s},q_{o}))  \nonumber
\end{align}
where $t(x,y)$ extracts the syntactic tree from text $x$, enriching it with REL tags computed with respect to $y$.
Thus $t$ is an asymmetric function. 

For task B we used the PTK kernel. Since the trees of the Arabic data are rather 
large and very noisy, we used STK, which is faster and uses less features. 

\subsection{Feature Vectors} 
\label{sec:proposedapproachesf}

Our L2R approach relies on various subsets of features to derive the relationship between two texts: text similarities, PTK similarity, Google rank, embedding features and machine translation evaluation features. 
In task B we used text similarities, PTK similarity, and Google rank. 
In task D we use text embedding features and machine translation evaluation features. 

\textbf{Text Similarity Features.}
We compute a total of 20 similarities, $sim(q_{o},q_j)$, using word $n$-grams ($n=[1,\ldots,4]$), after stopword removal, using greedy string tiling~\cite{Wise:1996}, longest common subsequences~\cite{Allison:1986}, Jaccard coefficient~\cite{Jaccard:1901}, word containment~\cite{Lyon:2001}, and cosine similarity.
%

\textbf{PTK Features.}
Another similarity is obtained by comparing syntactic trees with PTK, i.e., $TK(t(q_{o},q^{i}_{s}),t(q^{i}_{s},q_{o}))$. Note that, different from the model in Section~\ref{TK}, PTK here is applied to the members of the same pair and thus only produces one feature.

\textbf{Ranking-based Features.}
Our ranking feature is based on the ranking generated by Google. Each forum question is located in one position in the range $[1,\ldots,10]$. We try to exploit this information in two ways ``as-is'' ($pos$) or the inverse ($pos^{-1}$).%

\textbf{Embeddings}
We utilize the embedding vectors as obtained 
by~\cite{belinkov-EtAl:2015:SemEval}: employing word2vec~\cite{Mikolov:2013} on 
the Arabic Gigaword corpus~\cite{Parker2011}. More specifically, we concatenate 
the vectors representing a new question and an existing question in the 
question--answer pair, which is then fed to the SVM classifier. 

\textbf{MTE features}
We used machine translation evaluation features: BLEU \cite{Papineni:Roukos:Ward:Zhu:2002}, TER \cite{Snover06astudy}, Meteor \cite{Lavie:2009:MMA}, NIST \cite{Doddington:2002:AEM}, Precision and Recall, and length ratio between the question and the comment. 

\section{Experiments} 
\label{sec:results}

We follow the evaluation framework of SemEval 2016 Task 3 in order to be able 
to compare our system with the ones of the 
competition~\cite{nakov-EtAl:2016:SemEval}. 

We used different tools for preprocessing the text in English: OpenNLP's 
tokenizer, POS-tagger and chunk 
annotator\footnote{\url{https://opennlp.apache.org/}}, and Stanford's 
lemmatizer~\cite{manning-EtAl:2014:P14-5}, all accessible through DKPro 
Core~\cite{eckartdecastilho-gurevych:2014:OIAF4HLT}\footnote{\url{
https://dkpro.github.io/dkpro-core/}}. 
For Arabic texts were first removed stop-words, keeping only content and Latin words. We used the MADAMIRA toolkit~\cite{PASHA14.593} for segmenting the texts. In order to split the texts into sentences, we used the Stanford splitter.\footnote{\url{http://stanfordnlp.github.io/CoreNLP}} For parsing Arabic texts into syntactic trees, we used the Berkeley parser~\cite{petrov-klein:2007:main}. 

\subsection{Experiments on Task B}

In a set of preliminary experiments, we first compared a reranker 
---SVMrank~\cite{Joachims:2002:OSE:775047.775067}--- with a standard binary 
SVM~\cite{Joachims:99}. 
As the results were comparable, we employed SVMs using the KeLP 
toolkit\footnote{\url{https://github.com/SAG-KeLP}}, which 
enables to combine our three subsets of features within different kernels; 
namely RBF for the similarity features, tree kernels for the parse trees, and 
either linear or RBF kernels for the ranking-based feature. 
We set the C parameter of the SVMs to $1$ in all the experiments and the parameters of the TKs and RBF kernels to default values. 

We conducted three experiments with growing complexity for assessing the effectiveness of our different feature sets (see Section~\ref{sec:features}), with respect to GR. 
In agreement, with the SemEval challenge, we evaluate our rankings with Mean Average Precision (MAP), average Recall (AvgRec), and Mean Reciprocal Rank (MRR). 

We tested the performance of each of the feature sets in isolation and pair-wise. Table~\ref{tab:results-two-features} reports the obtained performance both on the development and test sets. 
The baseline is computed on GR, which produces a strong value as it is a product 
of the Google technology and its associated knowledge bases. We have two 
advantages, though: 
\Ni Google is not tuned up on the specific forum data we use and
\Nii probably Google  does not use syntactic structures in algorithms such as 
TKs.

Our results support the hypotheses above. Indeed, the MAP of the models derived 
by Similarities, TK, and their combinations is below GR: without accessing to 
the Google resources, our models can just approach the search engine's 
performance. However, when using the ranking feature, our best model 
outperforms the MAP of GR by $2.30$ and $1.66$ absolute percent points on the 
development and test sets, respectively. 
The RBF kernel on the ranking feature produces a larger improvement as it can 
more effectively express higher similarity values when the rankings of 
questions are close. This cannot be done  with a linear kernel. 

To better study the result above, Table~\ref{tab:results-three-feats} reports 
the combinations between the kernel function (Linear or RBF) and the 
representation of the ranking feature (the ranking itself or its inverse). 
While all combinations improve above the baseline, there is no clear indication 
on the choice between pos or pos$^{-1}$. However, the use of the RBF kernel 
results in the highest performance.

\begin{table}
\setlength{\tabcolsep}{3pt}
\centering
\caption{Ranking-based features combined with linear and RBF kernels on the 
Question--Question similarity dataset. $\dagger$ shows statistically significant 
results (at 95\%) with respect to GR.}
\footnotesize
\label{tab:results-two-features}
\begin{tabular}{l|ccc|ccc} 
						&\multicolumn{3}{c}{DEV}	& \multicolumn{3}{|c}{TEST}	\\\hline
Model		  			& MAP 	& AvgRec & MRR		& MAP 	& AvgRec & MRR	\\ \hline
GR baseline  	  			& 71.35$\dagger$	& 86.11	& 76.67$\dagger$ 		& 74.75$\dagger$ 	& 88.30 	& 83.79 \\\hline
Sim.						& 64.80	& 82.52	& 73.73		& 70.70	& 85.78	& 80.58	\\ 
TK						& 69.97	& 86.86	& 77.73	  	& 73.98	& 88.90	& 82.55 \\ 
TK + Sim 					& 71.07 	& 87.72 	& 78.14 		& 73.81	& 89.21	& 82.86 \\ \hline

\bf Linear	Kernel		&&&&&&\\
Sim + $pos$				& 68.04	& 85.07	& 76.00		& 71.99	& 87.92	& 81.19\\
Sim + $pos^{-1}$			& 70.17	& 85.98	& 78.17		& 75.15	& 89.19	& 84.29\\  
TK $\,$+ $pos$				& 71.77	& 88.46	& 78.12		& 75.34	& 90.67	& 83.19\\ 
TK $\,$+ $pos^{-1}$			& 72.64	& 87.69	& 75.58 		& 76.18$\dagger$	& 90.62	& 84.62 \\ \hline

\bf RBF	Kernel			&&&&&&\\
Sim. + $pos$				& 70.42	& 86.38	& 78.50		& 74.61	& 89.10	& 83.81	\\
Sim. + $pos^{-1}$			& 69.82	& 85.91	& 77.17		& 74.58	& 89.09	& 83.57\\ 
TK $\,$+ $pos$ 			& 72.93	& 87.95	& 77.54		& 75.72	& 90.80	& 83.86	\\ 
TK $\,$+ $pos^{-1}$ 			& 73.65$\dagger$	& 88.78		& 79.58$\dagger$ 	& 76.41$\dagger$ & 91.14 & 84.62 \\  
\hline
\end{tabular}
\vspace{-1em}
\end{table}


\begin{table}
\setlength{\tabcolsep}{3pt}
\centering
\caption{Performance of different rank features (all models include RBF kernel 
on similarities and tree kernel) on the Question--Question similarity dataset. 
$\dagger$ shows statistically significant results (at 95\%) with respect to GR.}
\label{tab:results-three-feats}\vspace{.2em}
\footnotesize
\begin{tabular}{l|ccc|ccc} 
						& \multicolumn{3}{c}{DEV}	& \multicolumn{3}{|c}{TEST}	\\\hline
Model		  			& MAP 	& AvgRec & MRR		& MAP 	& AvgRec	& MRR	\\ \hline
GR baseline  	  		& 71.35$\dagger$	& 86.11	& 76.67			& 74.75$\dagger$ & 88.30 & 83.79 \\
\bf Linear	Kernel		&&&&&&\\
$pos$  					& 72.18 & 88.41 & 78.00			& 75.67 & 90.77 & 83.38  \\  
$pos^{-1}$ 				& 73.28$\dagger$ & 88.47 & 80.00 			& 76.28$\dagger$ & 90.72 & 84.62 \\
\hline
\bf RBF	Kernel			&&&&&&\\
$pos$ 					& 73.24$\dagger$ & 88.37 & 78.40  			& 76.47$\dagger$ & 90.78 & 84.21  \\
$pos^{-1}$ 				& 73.60$\dagger$ & 88.85 & 79.67 			& 75.89 & 90.57 & 84.14 \\
\hline
\end{tabular}
\vspace{-1em}
\end{table}

UH-PRHLT-primary, the best system at competition 
time~\cite{nakov-EtAl:2016:SemEval}, obtained a MAP of 76.70. This value is not 
statistically different from our 76.47; still that system uses knowledge bases, 
such as BabelNet and FrameNet~\cite{Baker:1998:BFP:980845.980860}, which we do 
not require. 

\subsection{Experiments on Task D}

We have performed three sets of experiments for task D, which coincide with our submissions at the competition and which are reported in Table~\ref{tab:results}. 
In the first experiment, cont$_{1}$, we applied an SVM with a linear kernel on the embedding features. 
A second experiment, cont$_{2}$, also includes machine translation evaluation 
features (both features are described in 
Section~\ref{sec:proposedapproachesf}). 
Finally, a third experiment, {\it primary}, used a linear kernel on the
embedding features in combination with  the syntactic tree kernel 
(STK)~\cite{Moschitti:2006}, applied as described in 
Section~\ref{sec:proposedapproaches}, to the constituency trees of the question 
texts. 


\begin{table}[t]
\centering
\begin{tabular}{l|@{\hspace{2mm}}c@{\hspace{2mm}}c@{\hspace{2mm}}c@{\hspace{2mm
}}c}
    \bf Task D	     & MAP     &  AvgRec &  MRR & Rank\\ \hline 
%
  primary& $45.50$ & $50.13$ & $52.55$ & 2\\
  cont$_1$   & $38.33$ & $42.09$ & $43.75$ & 11 \\
  cont$_2$   & $39.98$ & $43.68$ & $46.41$ & 7 \\
  \hline
  best        & $45.83$ & $51.01$ & $53.66$ & 1 \\
  baseline   & $28.88$ & $28.71$ & $30.93$ & - \\
  \hline
 \end{tabular}
 \vspace{1em}
\caption{Performance of our primary and contrastive submissions to SemEval-2016 
Task D. Best-performing and baseline systems included for comparison. Rank 
stands  for the position in the challenge ranking. Baselines as provided 
reported in~\cite{nakov-EtAl:2016:SemEval}.
}
\label{tab:results}
\end{table}

%
%
%
The submission cont$_{1}$, using embedding features from \cite{belinkov-EtAl:2015:SemEval}, is an average system. When we add the machine translation evaluation (MTE) features the MAP increases from 38.33 to 39.98, making us jump from the 11th to the 7th position in the competition. 
However, when we combine tree kernels with embedding features MAP improves 
by more than 7 absolute points (45.50), achieving the second position in the competition, very close to the best system, i.e. 45.83. 

The results on the Arabic dataset give more evidence that exploiting syntactical 
information with tree kernels is crucial for achieving state-of-the-art 
performance and that tree kernels can be effective for different languages. 


\section{Conclusions}
\label{sec:conclusions}
We have tackled the task of question re-ranking in English and Arabic using 
different sets of features and exploiting the syntactical structure of the text 
via tree kernels. 
We showed that the combination of similarity features, syntactic structures 
based on tree kernels and features based on the ranking of search engines is 
able to boost the  performance of a question re-ranker on a real-world cQA 
dataset. 
In particular, our results suggest that Google uses general models that can be 
on par with specific models trained on specific domains. However, if we also use 
advanced syntactic/semantic representations for modeling the structural 
relations between questions, we can achieve better results. 
Moreover, we modeled and tested relational tree kernels for cQA, which are robust to noise and can thus boost Google's ranking. 
In the future, we would like to better structure the representation of the questions. Indeed, as mentioned before, there are several different sections of the question text, e.g., subquestions, subject, elaborations. These could be used to improve our shallow representation, which, at the moment, merges all the question trees in a flat macro-tree. 

In the comment reranking task we used for the first time tree kernels on an 
Arabic dataset and showed that they are a key component for reaching 
a performance comparable to the best SemEval-2016 system. 
In the future we plan to combine our system with the best one of the competition 
to further improve our results. Furthermore, we plan to address the challenges 
caused by the large size of the trees in order to be able to use more powerful 
tree kernels. 

\section*{Acknowledgments}
This research is developed by the Arabic Language Technologies (ALT) group at the Qatar Computing Research Institute (QCRI), HBKU, Qatar Foundation in collaboration with MIT. It is part of the Interactive sYstems for Answer Search ({\sc Iyas}) project. 

\bibliographystyle{abbrv}
\bibliography{sigproc}

%
%
%
%


\end{document}

%% file: abstractAndIntro.tex
\begin{abstract}
This paper studies the impact of different types of features applied to 
learning to re-rank questions in community Question Answering. We tested our 
models on two datasets released in SemEval-2016 Task 3 on ``Community Question 
Answering''. Task 3 targeted real-life Web fora both in English and Arabic. 
Our models include bag-of-words features (BoW), syntactic tree kernels (TKs), 
rank features, embeddings, and machine translation evaluation features. To the 
best of our knowledge, structural kernels have barely been applied to the 
question reranking task, where they have 
to model paraphrase relations. 
In the case of the English question re-ranking task, we compare our 
learning to rank (L2R) algorithms against a strong baseline given by the 
Google-generated ranking (GR). The results show that
\Ni the shallow structures used in our TKs are robust enough to noisy data and 
\Nii improving GR is possible, but effective BoW features and TKs along 
with an accurate model of GR features in the used L2R algorithm are required. 
In the case of the Arabic question re-ranking task, for the first time we 
applied tree kernels on syntactic trees of Arabic sentences. Our approaches to 
both tasks obtained the second best results on SemEval-2016 subtasks B on 
English and D on Arabic. 
\end{abstract}


\section{Introduction}
\label{sec:intro}

In recent years, there has been a renewed interest in information retrieval  
for community question answering\footnote{Commercial applications are currently using advanced technology such as the one we present in this paper: http://iyas.qcri.org/ql-demo} (cQA). 
This combines traditional question 
answering with a modern Web scenario, where users pose questions on a forum 
expecting to receive answers from other users. As fora are fairly open, many 
contributors may post comments loosely connected to the original question, 
whereas others are simply wrong or are not answers at all. We addressed this 
problem in~\cite{barroncedeno-EtAl:2015:ACL-IJCNLP, 
joty-EtAl:2015:EMNLP1,nicosia-EtAl:2015:SemEval}. 

One of the most critical problem arises when a user posts a new question to the 
forum. In this case, the retrieval system would search for relevant comments 
linked to other questions in order to find potentially appropriate answers. In 
this noisy and complex setting, also powerful search engines (e.g., Google), 
have hard time to retrieve comments that can correctly answer the original 
question. An approach to deal with this case is to divide the problem into two 
separate tasks: 
\Ni retrieving a set of similar questions and their associated set of 
comments and 
\Nii assessing the usefulness of the retrieved comments with respect to the 
question posed by the user. In this paper we focus on first task both in 
English and Arabic. This shows that our models can be applied to different 
languages.

The retrieval of relevant questions requires models different from 
typical search engines. Firstly, we deal with short texts: questions may include 
descriptions or subquestions but they are not typically larger than one or two 
paragraphs. Thus rich similarity features are necessary to deal effectively 
with the sparseness of such texts.
Secondly, the syntactic structure of the questions is rather important for 
learning to recognize paraphrases and thus selecting the right candidates. For 
instance, consider the following two questions :

\begin{description}
\item[$q_1$] How do I get a visa for Qatar 
to visit my wife?
\item[$q_2$] How do I get a visa for my wife to have her visit Qatar?
\end{description}
Question $q_1$ has roughly the same BoW representation as question $q_2$,
but their information request ---therefore their answers--- are totally 
different. The structure of the questions can be exploited to detect the 
difference between these difficult cases. 


In the past, the study of cQA models has been limited by a substantial lack of 
manually annotated data for question--question similarity, as the main 
annotation was provided by the user themselves and thus essentially not fully 
reliable. Recently, a new resource has been released for the SemEval 2016 Task 3 
on Answer Selection in cQA~\cite{nakov-EtAl:2016:SemEval}.%
\footnote{\url{http://alt.qcri.org/semeval2016/task3}}  
Given a set of existing forum questions $Q$, where each existing question $q \in 
Q$ is associated with a set of answers $C_q$, and a new user question $q'$, the 
ultimate task is to determine whether a comment $c \in C_q$ represents a 
pertinent answer to $q'$ or not. This task can be subdivided into three tasks, 
namely: 
\Na to assign a relevance (\emph{goodness}) score to each answer $c \in C_q$ with respect to the existing question $q$; \Nb to re-rank the set of questions $Q$ according to their relevance against the new question $q'$; and finally  
\Nc to predict the appropriateness of the answers $c \in C_q$ against $q'$. 
An adaptation of Task C was proposed for Arabic (Task D). 
In this paper, we focus on tasks B and D: question re-ranking in both English 
and Arabic. 

For task B we use 
\Ni text similarity features, derived from the classical BoW representations, 
e.g., n-grams, skip-grams; 
\Nii syntactic/structural features injected by TKs, which have shown to achieve 
the state of the art in the related task of answer sentence 
retrieval~\cite{Tymoshenko:2015:AIS:2806416.2806490}; and 
\Niii features for modeling the initial rank provided by a state-of-the-art 
search engine, which represents a strong baseline. 
Our extensive experimentation, 
produced the following results: 
\Ni the BoW features based on similarity measures alone do not improve GR. 
\Nii Our TKs  applied to questions alone outperform the models based on 
similarity measures and when jointly used with the rank features improve all the 
models. 
In particular, they outperform GR by 1.72 MAP points (95\% of statistical 
confidence). 

For task D we apply tree kernels on pairs of syntactic trees of Arabic 
sentences and define additional features derived from machine translation 
evaluation scores. 

Our approaches to both tasks B and D granted us the second position on SemEval 
2016 Task 3. 
\medskip

The rest of our contribution is distributed as follows. 
Section~\ref{sec:related} includes related work. 
Section~\ref{sec:taskdescription} describes the approached problem. 
Section~\ref{sec:features} describes our learning to rank models. 
Section~\ref{sec:results} discusses our experiments and obtained results.
Section~\ref{sec:conclusions} includes conclusions and further work.

%% file: 2016_qcri_semeval3.bbl
\begin{thebibliography}{10}

\bibitem{Allison:1986}
L.~Allison and T.~Dix.
\newblock A bit-string longest-common-subsequence algorithm.
\newblock {\em Inf. Process. Lett.}, 23(6):305--310, Dec. 1986.

\bibitem{Baker:1998:BFP:980845.980860}
C.~F. Baker, C.~J. Fillmore, and J.~B. Lowe.
\newblock The berkeley framenet project.
\newblock In {\em Proceedings of the 36th Annual Meeting of the Association for
  Computational Linguistics and 17th International Conference on Computational
  Linguistics - Volume 1}, ACL '98, pages 86--90, Stroudsburg, PA, USA, 1998.
  Association for Computational Linguistics.

\bibitem{barroncedeno-EtAl:2015:ACL-IJCNLP}
A.~Barr\'{o}n-Cede\~{n}o, S.~Filice, G.~Da~San~Martino, S.~Joty,
  L.~M\`{a}rquez, P.~Nakov, and A.~Moschitti.
\newblock Thread-level information for comment classification in community
  question answering.
\newblock In {\em Proceedings of the 53rd Annual Meeting of the Association for
  Computational Linguistics and the 7th International Joint Conference on
  Natural Language Processing (Volume 2: Short Papers)}, pages 687--693,
  Beijing, China, July 2015. Association for Computational Linguistics.

\bibitem{belinkov-EtAl:2015:SemEval}
Y.~Belinkov, M.~Mohtarami, S.~Cyphers, and J.~Glass.
\newblock {VectorSLU}: A continuous word vector approach to answer selection in
  community question answering systems.
\newblock In {\em Proceedings of the 9th International Workshop on Semantic
  Evaluation}, SemEval~'15, Denver, Colorado, USA, 2015.

\bibitem{cao2009use}
X.~Cao, G.~Cong, B.~Cui, C.~S. Jensen, and C.~Zhang.
\newblock The use of categorization information in language models for question
  retrieval.
\newblock In {\em CIKM}, pages 265--274, 2009.

\bibitem{Doddington:2002:AEM}
G.~Doddington.
\newblock Automatic evaluation of machine translation quality using n-gram
  co-occurrence statistics.
\newblock In {\em Proceedings of the Second International Conference on Human
  Language Technology Research}, HLT '02, pages 138--145, 2002.

\bibitem{duan2008searching}
H.~Duan, Y.~Cao, C.-Y. Lin, and Y.~Yu.
\newblock Searching questions by identifying question topic and question focus.
\newblock In {\em ACL}, pages 156--164, 2008.

\bibitem{eckartdecastilho-gurevych:2014:OIAF4HLT}
R.~Eckart~de Castilho and I.~Gurevych.
\newblock A broad-coverage collection of portable nlp components for building
  shareable analysis pipelines.
\newblock In {\em Proceedings of the Workshop on Open Infrastructures and
  Analysis Frameworks for HLT}, Dublin, Ireland, August 2014.

\bibitem{Jaccard:1901}
P.~Jaccard.
\newblock {\'{E}tude comparative de la distribution florale dans une portion
  des Alpes et des Jura}.
\newblock {\em Bulletin del la Soci\'{e}t\'{e} Vaudoise des Sciences
  Naturelles}, pages 547--579, 1901.

\bibitem{ji2012question}
Z.~Ji, F.~Xu, B.~Wang, and B.~He.
\newblock Question-answer topic model for question retrieval in community
  question answering.
\newblock In {\em CIKM}, pages 2471--2474, 2012.

\bibitem{Joachims:99}
T.~Joachims.
\newblock {Making Large-scale Support Vector Machine Learning Practical}.
\newblock In {\em {Advances in Kernel Methods}}. {MIT Press}, {Cambridge, MA,
  USA}, 1999.

\bibitem{Joachims:2002:OSE:775047.775067}
T.~Joachims.
\newblock Optimizing search engines using clickthrough data.
\newblock KDD, pages 133--142, 2002.

\bibitem{joty-EtAl:2015:EMNLP1}
S.~Joty, A.~Barr\'{o}n-Cede\~{n}o, G.~Da~San~Martino, S.~Filice,
  L.~M\`{a}rquez, A.~Moschitti, and P.~Nakov.
\newblock Global thread-level inference for comment classification in community
  question answering.
\newblock In {\em Proceedings of the 2015 Conference on Empirical Methods in
  Natural Language Processing}, pages 573--578, Lisbon, Portugal, September
  2015. Association for Computational Linguistics.

\bibitem{Lavie:2009:MMA}
A.~Lavie and M.~Denkowski.
\newblock The {METEOR} metric for automatic evaluation of machine translation.
\newblock {\em Machine Translation}, 23(2--3):105--115, 2009.

\bibitem{Lyon:2001}
C.~Lyon, J.~Malcolm, and B.~Dickerson.
\newblock Detecting short passages of similar text in large document
  collections.
\newblock EMNLP, pages 118--125, 2001.

\bibitem{manning-EtAl:2014:P14-5}
C.~D. Manning, M.~Surdeanu, J.~Bauer, J.~Finkel, S.~J. Bethard, and
  D.~McClosky.
\newblock The {Stanford} {CoreNLP} natural language processing toolkit.
\newblock In {\em Association for Computational Linguistics (ACL) System
  Demonstrations}, pages 55--60, 2014.

\bibitem{Mikolov:2013}
T.~Mikolov, W.-t. Yih, and G.~Zweig.
\newblock {Linguistic Regularities in Continuous Space Word Representations}.
\newblock In {\em {Proceedings of the 2013 Conference of the North American
  Chapter of the Association for Computational Linguistics: Human Language
  Technologies}}, NAACL-HLT '13, pages 746--751, {Atlanta, GA, USA}, 2013.

\bibitem{Moschitti:2006}
A.~Moschitti.
\newblock {Efficient Convolution Kernels for Dependency and Constituent
  Syntactic Trees}.
\newblock In {\em {ECML}}, pages 318--329. 2006.

\bibitem{nakov-EtAl:2016:SemEval}
P.~Nakov, L.~M\`{a}rquez, A.~Moschitti, W.~Magdy, H.~Mubarak, A.~A. Freihat,
  J.~Glass, and B.~Randeree.
\newblock {SemEval}-2016 task 3: Community question answering.
\newblock In {\em Proceedings of SemEval '16}. ACL, 2016.

\bibitem{nicosia-EtAl:2015:SemEval}
M.~Nicosia, S.~Filice, A.~Barr\'{o}n-Cede\~{n}o, I.~Saleh, H.~Mubarak, W.~Gao,
  P.~Nakov, G.~Da~San~Martino, A.~Moschitti, K.~Darwish, L.~M\`{a}rquez,
  S.~Joty, and W.~Magdy.
\newblock {QCRI}: Answer selection for community question answering -
  experiments for {A}rabic and {E}nglish.
\newblock In {\em Proceedings of the 9th International Workshop on Semantic
  Evaluation}, SemEval~'15, Denver, Colorado, USA, 2015.

\bibitem{Papineni:Roukos:Ward:Zhu:2002}
K.~Papineni, S.~Roukos, T.~Ward, and W.-J. Zhu.
\newblock {BLEU}: a method for automatic evaluation of machine translation.
\newblock In {\em Proceedings of 40th Annual Meting of the Association for
  Computational Linguistics}, ACL '02, pages 311--318, Philadelphia,
  Pennsylvania, USA, 2002.

\bibitem{Parker2011}
R.~Parker, D.~Graff, K.~Chen, J.~Kong, and K.~Maeda.
\newblock {\em Arabic Gigaword Fifth Edition}.
\newblock Linguistic Data Consortium (LDC), Philadelphia, 2011.

\bibitem{PASHA14.593}
A.~Pasha, M.~Al-Badrashiny, M.~Diab, A.~E. Kholy, R.~Eskander, N.~Habash,
  M.~Pooleery, O.~Rambow, and R.~Roth.
\newblock Madamira: A fast, comprehensive tool for morphological analysis and
  disambiguation of arabic.
\newblock In {\em Proceedings of the Ninth International Conference on Language
  Resources and Evaluation (LREC'14)}, Reykjavik, Iceland, may 2014.

\bibitem{petrov-klein:2007:main}
S.~Petrov and D.~Klein.
\newblock Improved inference for unlexicalized parsing.
\newblock In {\em Human Language Technologies 2007: The Conference of the North
  American Chapter of the Association for Computational Linguistics;
  Proceedings of the Main Conference}, pages 404--411, Rochester, New York,
  April 2007. Association for Computational Linguistics.

\bibitem{sigir12}
A.~Severyn and A.~Moschitti.
\newblock Structural relationships for large-scale learning of answer
  re-ranking.
\newblock SIGIR, pages 741--750, 2012.

\bibitem{Snover06astudy}
M.~Snover, B.~Dorr, R.~Schwartz, L.~Micciulla, and J.~Makhoul.
\newblock A study of translation edit rate with targeted human annotation.
\newblock In {\em Proceedings of the 7th Biennial Conference of the Association
  for Machine Translation in the Americas}, AMTA '06, Cambridge, Massachusetts,
  USA, 2006.

\bibitem{Tymoshenko:2015:AIS:2806416.2806490}
K.~Tymoshenko and A.~Moschitti.
\newblock Assessing the impact of syntactic and semantic structures for answer
  passages reranking.
\newblock In {\em Proceedings of CIKM '15}, pages 1451--1460, New York, NY,
  USA, 2015. ACM.

\bibitem{wang2009syntactic}
K.~Wang, Z.~Ming, and T.-S. Chua.
\newblock A syntactic tree matching approach to finding similar questions in
  community-based qa services.
\newblock In {\em SIGIR}, pages 187--194, 2009.

\bibitem{Wise:1996}
M.~Wise.
\newblock Yap3: Improved detection of similarities in computer program and
  other texts.
\newblock In {\em SIGCSE}, pages 130--134, 1996.

\bibitem{zhang2014question}
K.~Zhang, W.~Wu, H.~Wu, Z.~Li, and M.~Zhou.
\newblock Question retrieval with high quality answers in community question
  answering.
\newblock In {\em CIKM}, pages 371--380, 2014.

\bibitem{zhou2011phrase}
G.~Zhou, L.~Cai, J.~Zhao, and K.~Liu.
\newblock Phrase-based translation model for question retrieval in community
  question answer archives.
\newblock In {\em ACL}, pages 653--662, 2011.

\end{thebibliography}
